\documentclass{article}
\usepackage[preprint]{neurips_2026}

\usepackage[utf8]{inputenc} 
\usepackage[T1]{fontenc}    
\usepackage{hyperref}       
\usepackage{url}            
\usepackage{booktabs}       
\usepackage{amsfonts}       
\usepackage{nicefrac}       
\usepackage{microtype}      
\usepackage[table]{xcolor}
\usepackage{graphicx}
\usepackage{amsmath,amssymb}
\usepackage{annotate_equations}
\usepackage{wrapfig}
\usepackage{makecell}
\definecolor{col1}{RGB}{232, 161, 148}
\definecolor{col2}{RGB}{148, 187, 232}
\usepackage{algorithm}
\usepackage{algorithmicx}
\usepackage{amsthm}
\usepackage{algpseudocode}
\usepackage{multirow}
\usepackage{capt-of}

\newcommand{\method}{Stream3D}

\usepackage{amsmath,amsfonts,bm}

\def\eqref#1{equation~\ref{#1}}

\def\1{\bm{1}}

\DeclareMathAlphabet{\mathsfit}{\encodingdefault}{\sfdefault}{m}{sl}
\SetMathAlphabet{\mathsfit}{bold}{\encodingdefault}{\sfdefault}{bx}{n}

\title{Stream3D: Sequential Multi-View 3D Generation \\ via Evidential Memory}

\author{%
  Kaichen Zhou$^{2,3*}$ \ \ Zeyang Bai$^{1,2*}$ \ \ Xinhai Chang$^{2*}$  \\
  \textbf{Mengyu Wang}$^{3\dagger}$ \ \ \textbf{Paul Pu Liang}$^{2\dagger}$ \ \ \textbf{Fangneng Zhan}$^{1\dagger}$ \\ \\
  $^{1}$World Mind Lab, HKUST \ \  $^{2}$Media Lab and EECS, MIT \\
  $^{3}$Kempner Institute, Harvard University \\
}

\begin{document}

\renewcommand{\thefootnote}{}
\footnotetext{$^*$Equal contribution as first authors. $^\dagger$Joint supervision.}

\maketitle

\begin{center}
›\includegraphics[width=\linewidth]{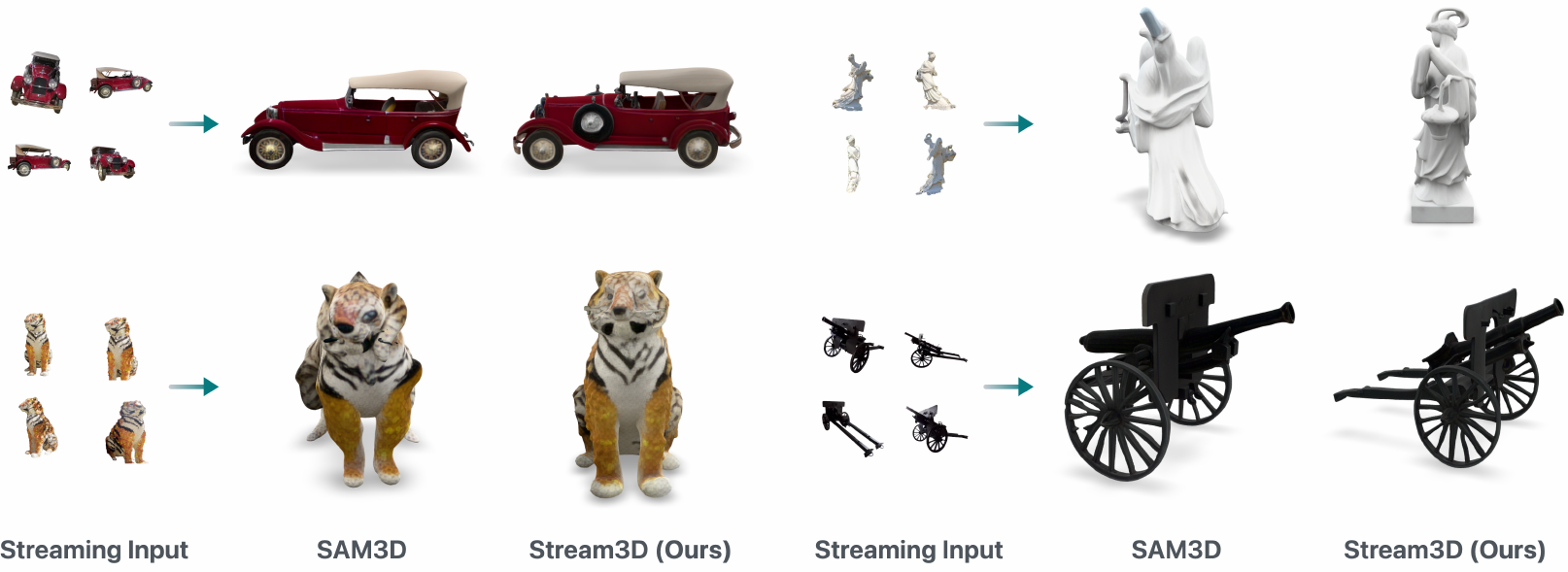}
\end{center}
\vspace{-0.2cm}
\captionof{figure}{
  \textbf{\method{}} takes streaming input views as additional conditioning signals to improve the performance of pretrained single-view-conditioned 3D generation models without retraining pretrained weights. Compared with SAM-3D, incorporating views from the input stream can substantially improve 3D generation quality.}
  \label{fig:teaser}
\vspace{0.3cm}

\begin{abstract}
View-conditioned 3D generators such as SAM~3D, TRELLIS and Hunyuan3D produce high-quality object 3D representations from a single view, but real-world visual observation often arrives as long monocular streams.
Naively applying these generators to each streaming frame independently leads to severe temporal inconsistency in the generated results.
To address this problem, we propose \method{}, the first training-free streaming mechanism that turns a frozen view-conditioned 3D generator into a streaming generator with constant cross-chunk memory. 
\method{} achieves this by maintaining a compact evidential memory, which selectively caches the most informative historical frames based on a proposed evidence score mechanism. As the stream progresses, the memory dynamically updates to retain a fixed number of informative frames, preventing the memory footprint from growing linearly with sequence length. This also prevents degradation over long sequences and keeps the underlying generator completely unchanged without retraining, architectural modifications, or auxiliary losses.
Evaluated on both realistic and synthetic streaming benchmarks, \method{} outperforms latent-transport baselines, including KV-cache reuse and flow-based feature editing, across both photometric and geometric metrics. 
More details can be found at: 
\href{https://stream-3d.github.io/stream3d.github.io/}{\nolinkurl{Link}}.
\end{abstract}

\section{Introduction}

Object-centric 3D generation is becoming a practical building block for vision and robotics~\cite{liu2023one, li2023instant3d, tang2024lgm}. 
Recent systems such as SAM~3D~Objects~\cite{chen2025sam} and TRELLIS.2~\cite{trellis2} can take an image and reconstruct corresponding Gaussian splat~\cite{kerbl20233d} or 3D mesh within seconds. 
However real capture devices produce long monocular streams: a phone circling an object~\cite{nerf} or a robot observing a scene while moving~\cite{photoslam}. 
Naively applying a single-view 3D generator frame by frame yields temporally inconsistent reconstructions and exploits only partial observations~\cite{chen2025sam,trellis2}.
Feeding all frames jointly through multi-diffusion~\cite{bar2023multidiffusion} or multiview fusion~\cite{li2026mv} is computationally expensive and can become infeasible for long streams, while processing fixed-size chunks with flow-matching-based editing~\cite{kulikov2025flowedit} discards historical context needed for global consistency.

An alternative approach is to adapt techniques from streaming video generation or 3D reconstruction~\cite{fifodiffusion,streamingt2v,lan2025stream3r,zhuo2025streaming}. For instance, state-transport mechanisms like KV banks~\cite{lan2025stream3r} or FlowEdit-style velocity edits~\cite{kulikov2025flowedit} could be introduced to propagate information across chunks. 
However, transporting latent states in this manner is also problematic. 
As the orientation, shape, and scale of these states are deeply entangled within the generator, they are difficult to align across frames, leading to severe error accumulation during the streaming process~\cite{li2026mv}.
Furthermore, as these methods transport and accumulate state over time, their memory footprint naturally grows with sequence length~\cite{chen2026longlive}. 
Such state propagation may also introduce compounding errors over long streams, echoing the error accumulation and memory bottlenecks observed in autoregressive video diffusion models~\cite{wang2025error}. 
Thus, the mechanism intended to preserve consistency can become a source of degradation when the transported state is misaligned or stale.

To solve this problem, we propose \textbf{\method}, a training-free streaming mechanism that turns a frozen view-conditioned 3D generator into a long-stream generator without modifying its weights or architecture. 
Rather than transporting latent states across chunks, \method{} retains \emph{evidential views}: observed frames that the generator itself identifies as reliable conditioning signals through its cross-attention maps. 
Specifically, during a lightweight warmup pass, we compute an evidence score for each query token and incoming views, measuring whether the token attends to that views both strongly and selectively. 
These scores update a token-level \emph{Adaptive Evidential Memory}, which stores only a fixed number of high-evidence frame indices per query token, keeping memory constant with stream length. 
This token-level design allows different spatial regions of the 3D volume to retrieve different historical observations, rather than relying on a single global frame subset or an accumulated latent state. 
As new views arrive, the retained evidence for each query token can only be maintained or improved, yielding a simple non-degradation property in evidence space. For generation, token-level evidence is aggregated into frame-level ownership scores, and the top-$K$ frames are selected as a bounded conditioning bundle for \emph{Evidence-Based Multi-Generation}. In this way, \method{} avoids latent-space alignment, preserves long-range visual evidence, and enables coherent streaming 3D generation while leaving the original generator unchanged.
We evaluate Stream3D as a lightweight mechanism mechanis around pre-trained SAM~3D on long monocular streams as shown in Fig.~\ref{fig:teaser}.
Our contributions are fourfold:
\begin{itemize}
\item \textbf{Streaming 3D generation.}
We pioneer the task of extending frozen view-conditioned 3D generators to long monocular streams, producing temporally consistent 3D generations while keeping memory bounded and avoiding retraining.
\item \textbf{Adaptive Evidential Memory.}
We introduce a compact memory mechanism that stores token-level evidential views rather than transporting latent states. The memory is constant and training-free, with a footprint that remains constant with stream length.
\item \textbf{Evidence-Based Multi-Generation.}
We propose an evidence-guided generation strategy that aggregates token-level memory into a bounded conditioning bundle and runs the frozen generator on the selected views. This enables different spatial regions to draw from the most reliable historical observations without modifying the underlying generator.
\item \textbf{Superior streaming performance.}
We show that \method{} outperforms latent-transport and fixed-view baselines across photometric and geometric metrics, while avoiding long-horizon degradation and maintaining a constant memory footprint.
\end{itemize}

\section{Related Work}
\subsection{3D Generation}
Recent object-centric 3D generators can be understood along three main design axes: \emph{output representation}, \emph{learning regime}, and \emph{input format}. Along the first axis, different methods adopt different 3D representations, including triplane- and Neural Radiance Field (NeRF)-based backbones \cite{lrm, triposr, nerf, pixelnerf, eg3d, li2023instant3d, openlrm, pflrm}, surface meshes \cite{instantmesh, unique3d, trellis, liu2023one, crm, meshlrm, makeit3d, get3d}, 3D Gaussian splats \cite{spar3d, kerbl20233d, tang2024lgm, grm, dreamgaussian, gaussiandreamer}, and native structured latent volumes \cite{hunyuan3d, direct3d, triposg, craftsman3d, chen2025sam, trellis2}. Along the second axis, existing approaches cover closed-form regressors \cite{lrm, triposr, instantmesh, unique3d, spar3d, trellis2, li2023instant3d, liu2023one, tang2024lgm, crm, meshlrm, openlrm, grm, pflrm}, 3D latent diffusion models \cite{hunyuan3d, direct3d, triposg, craftsman3d, chen2025sam, pointE, shapeE}, and SDS-based optimization methods \cite{dreamfusion, magic3d, prolificdreamer, sjc, latentnerf, fantasia3d, makeit3d, magic123, dreamgaussian, gaussiandreamer, dreamcraft3d, richdreamer}. Some layout-aware variants \cite{chen2025sam, li2026mv, renderdiffusion} further model the spatial arrangement of multiple objects within a scene.
The third axis, \emph{input format}, is the one most relevant to our work. Despite their differences in representation and learning paradigm, most existing methods assume a fixed and limited input interface: they take either a single image or a small, predefined set of images as input. Multi-view diffusion methods \cite{zero123, zero123pp, syncdreamer, wonder3d, mvdream, eschernet, sv3d, era3d} relax this setting by first hallucinating additional views before reconstructing 3D geometry. Video-conditioned reconstruction models \cite{dust3r, mast3r, fast3r, zhou2026page, vggt, longlrm} and 4D-aware generators \cite{fourdfy, mav3d, dreamin4d, animate124, consistent4d, mosaicv, vm4, zhou2026gem} extend the input format further to short temporal clips. More recently, MV-SAM3D~\cite{li2026mv} enables multi-view fusion directly in the latent space of a 3D generator. Nevertheless, these methods still operate on a fixed input bundle determined in advance, rather than supporting truly open-ended streaming inputs.
None of these methods natively supports unbounded online streams. In practice, long-stream processing is typically handled through ad hoc latent-transport schemes, which propagate intermediate states across chunks. However, such designs usually incur memory growth with sequence length and lead to performance degradation over long horizons.
%
In this work, we adapt a frozen view-conditioned generator \cite{chen2025sam} into a streaming generator that maintains constant cross-chunk memory, independent of stream length, and agnostic to the model backbone.

\subsection{Reconstruction from Streaming Inputs}
Reconstructing 3D geometry from streaming inputs has traditionally been studied in monocular SLAM~\cite{engel2014lsd,dso,davison2007monoslam,klein2007parallel,newcombe2011dtam,forster2014svo,mur2015orb,mur2017orb,orbslam3,neuralrecon,monogs,photoslam, zhou2025manydepth2}, which incrementally estimates camera motion and scene structure from video. Recent methods have extended modern feed-forward reconstruction models to online settings.
Notably, Spann3R~\cite{wang20253d} augments a DUSt3R-style encoder~\cite{dust3r} with a token-addressable spatial memory, enabling online pointmap fusion over long image streams. 
Similarly, SLAM3R~\cite{liu2025slam3r}, also built upon DUSt3R, introduces a real-time end-to-end dense reconstruction system that directly predicts 3D pointmaps from RGB videos. 
Point3R~\cite{wu2025point3r} further incorporates an explicit geometry-aligned spatial pointer memory, together with 3D hierarchical RoPE and an adaptive fusion mechanism. 
However, DUSt3R itself remains inherently two-view, restricting each inference step to a fixed image pair and making large-scale fusion dependent on iterative matching and optimization. 
VGGT-SLAM~\cite{maggio2025vggt,vggt} addresses this limitation by adopting the more powerful VGGT transformer, which supports image sets of arbitrary length.
CUT3R~\cite{wang2025continuous} instead adopts an RNN-style formulation for causal pointmap prediction from unstructured image streams. However, it compresses all past observations into a limited recurrent state, which can hinder long-range memorization and fine-grained multi-view fusion. 
Following the design philosophy of modern large language models, StreamVGGT~\cite{zhuo2025streaming} and Stream3R~\cite{lan2025stream3r} employ causal transformers to implicitly cache historical visual tokens. In addition, because CUT3R suffers from severe drift on long streaming inputs, TTT3R~\cite{chen2025ttt3r} proposes a simple empirical state-update rule to improve sequence-length generalization.
Stream3D departs from these reconstruction-centered approaches: while online reconstruction focuses on aggregating geometry already observed in the input stream, streaming 3D generation must also infer and synthesize unseen structure under temporal and geometric consistency constraints. 

\section{Method}
\label{sec:method}

\begin{figure*}[t]
\begin{center}
\includegraphics[width=\linewidth]{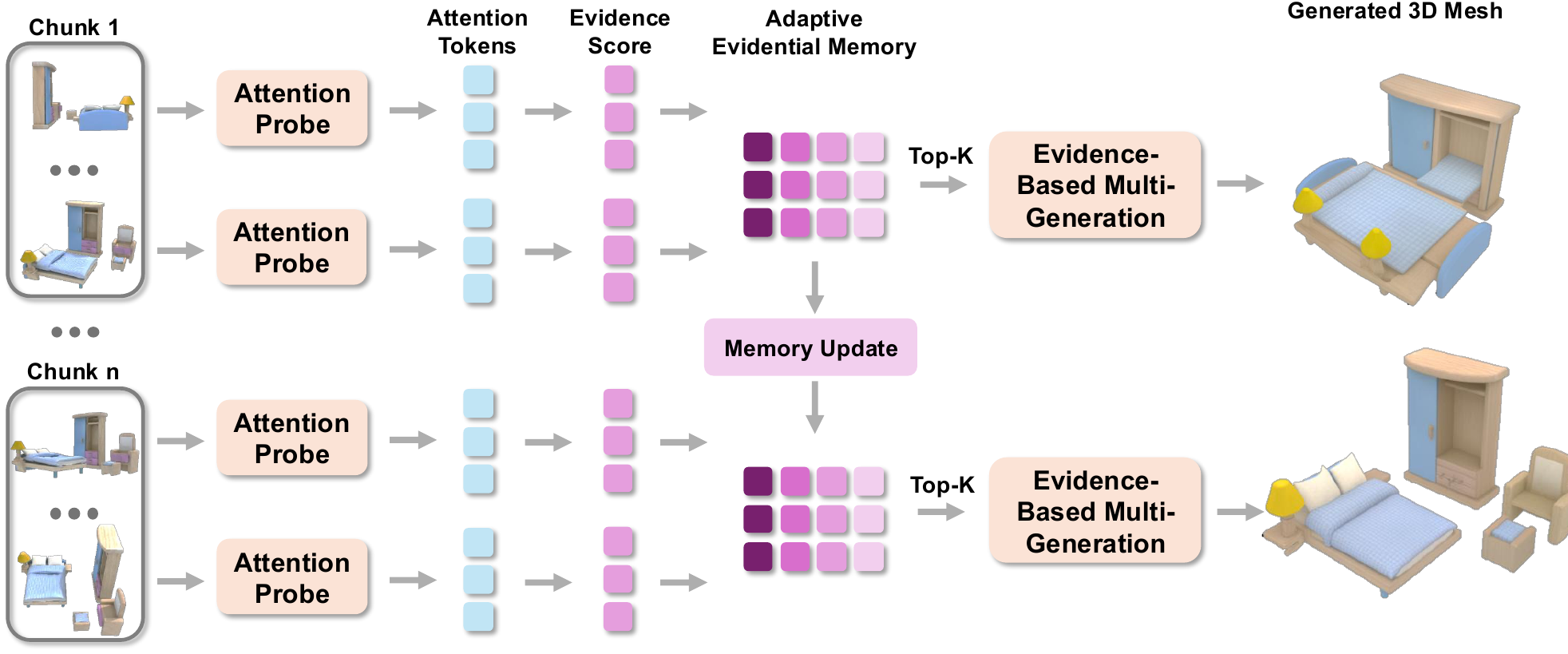}
\end{center}
\vspace{-0.5cm}
\caption{\textbf{Framework of \method{}.}
Given a streaming video, Stream3D processes frames chunk by chunk. A lightweight warmup pass extracts token-wise evidence score from cross-attention, which is stored to vote for informative frames to update the evidential memory, i.e., \textbf{Adaptive Evidential memory}.
Then, the top-K informative frames are passed to the \textbf{Evidence-Based Multi-Generation} for 3D asset generation.
By retaining only compact evidential memory rather than latent states, Stream3D achieves stable long-horizon generation with a constant memory footprint.
}
\label{fig:pipeline}
\vspace{-0.5cm}
\end{figure*}

We extend view-conditioned 3D generators, SAM~3D~Objects~\cite{chen2025sam} and TRELLIS.2~\cite{trellis2} --- to handle long streaming inputs without retraining shown in Fig.~\ref{fig:pipeline}. 
The \emph{Adaptive Evidential Memory} (Sec.~\ref{sec:method:cache}) enables efficient long-range memory by retaining token-level evidence from past views, while \emph{Evidence-Based Multi-Generation} (Sec.~\ref{sec:method:generation}) leverages this memory to produce temporally consistent and geometrically coherent 3D generations throughout the streaming process.

\subsection{Problem Setup}
\label{sec:method:setup}

\textbf{3D Generation Preliminary.}
Let $f_\theta$ denote a frozen view-conditioned 3D generator that, given a single input frame $v$ and an initial noise prior $z_0 \sim \mathcal{N}(0, I)$, produces a 3D sample (Gaussian splat~\cite{kerbl20233d}, mesh, or latent volume) $\hat{y} = f_\theta(v; z_0)$. Recent 3D generation models, i.e., SAM~3D~\cite{chen2025sam}, TRELLIS~\cite{trellis}, TRELLIS.2~\cite{trellis2}, Hunyuan3D 2.0~\cite{hunyuan3d} and CraftsMan3D~\cite{craftsman3d}, instantiate $f_\theta$ as a two-stage pipeline --- a structure stage (SS) producing a coarse occupancy / latent grid, followed by a texture or appearance stage (SLAT) --- with at least one cross-attention layer of the form:
\begin{figure}[h]
\vspace{1.5em}
\begin{equation}
\label{eq:xattn}
\eqnmarkbox[Emerald]{Av}{\mathbf{A}_v}
\;=\;
\mathrm{softmax}\!\left(
\frac{
\eqnmarkbox[NavyBlue]{Q}{\mathbf{Q}}\,
\eqnmarkbox[WildStrawberry]{Kattn}{\mathbf{K}^{\top}}
}{
\eqnmarkbox[BurntOrange]{sqrtd}{\sqrt{d}}
}
\right)
\;\in\;
\eqnmarkbox[Plum]{shape}{[0,1]^{Q \times P}},
\end{equation}
\annotate[yshift=1em]{above,left}{Av}{Cross-Attention Map}
\annotate[yshift=-2.5em]{below,left}{Q}{Query Tokens of Generator's Grid}
\annotate[yshift=1.5em]{above,right}{Kattn}{Key Tokens of Input Frame}
\annotate[yshift=-0.5em]{below,right}{sqrtd}{Scale Factor}
\annotate[yshift=1em]{above,right}{shape}{Attention Shape}
\vspace{0.1em}
\end{figure}

where $Q$ is the number of query tokens in the generator's voxel grid (e.g., $16^3 = 4096$ in SAM~3D's structure stage), $P$ is the number of key tokens in the input frame $v$, and $d$ is the feature dimension of each query/key token. 
In its native form, 3D generator $f_\theta$ accepts a single frame and produces a single reconstruction; it has no mechanism for incorporating multi-view evidence. 

\textbf{Streaming 3D Generation.}
Our method extends 3D generator $f_\theta$ to accept a stream of condition views $\mathcal{V} = \{v_1, \ldots, v_T\}$ by fusing the per-view forward passes through a confidence-weighted Multi-Diffusion-style aggregation in 3D latent space.
The stream is partitioned into overlapping chunks $\mathcal{C}_k$ of size $C$. At chunk $k$, we wish to produce a 3D sample $\hat{y}_k$ that is consistent with all previous chunks, while maintaining a memory that does not scale linearly with stream length $T$.

\subsection{Adaptive Evidential Memory} 
\label{sec:method:cache}

\begin{figure*}[t]
\begin{center}
\includegraphics[width=\linewidth]{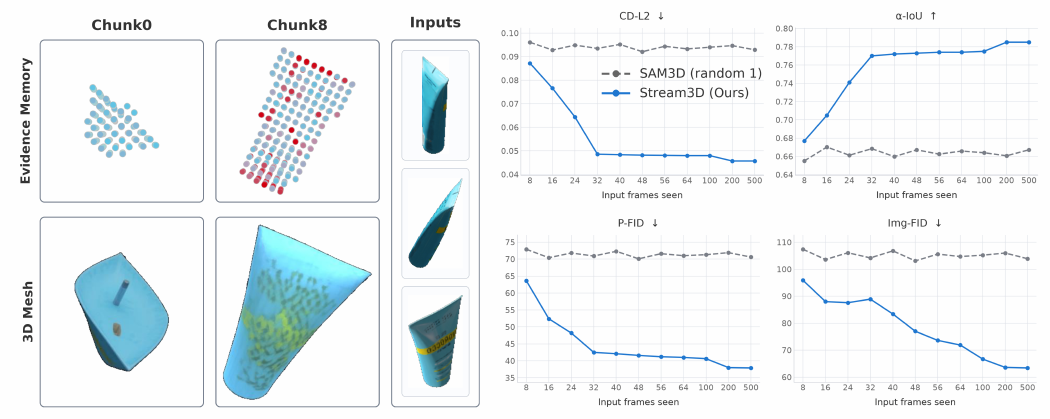}
\end{center}
\vspace{-0.5cm}
\caption{\textbf{Adaptive Evidential Memory.}
Left: Given a streaming sequence, our method updates a compact token-level memory by retaining historical views with the strongest evidence for each query token. The color transition from \textcolor{blue}{blue} to \textcolor{red}{red} denotes increasing frame indices, from earlier to later observations. Over time, the memory incorporates evidence from increasingly diverse viewpoints, and the reconstruction improves from early to later chunks, consistent with the long-horizon geometry and appearance curves. Right: Compared with SAM3D, Stream3D consistently improves performance as the number of observed chunks increases, and then reaches a stable plateau.}
\label{fig:memory}
\vspace{-0.5cm}
\end{figure*}

To maintain an efficient memory footprint that does not scale linearly with the total sequence length $T$ while preserving generation quality over long streams, we introduce an \emph{Adaptive Evidential Memory} mechanism, shown on the left of Fig.~\ref{fig:memory}. The right side of Fig.~\ref{fig:memory} shows that Stream3D improves as more frames are observed and then reaches a stable performance plateau.
Rather than blindly retaining all historical frames, this approach dynamically filters and preserves only the most informative views by evaluating their relevance at a per-token level. Our mechanism operates in three distinct stages. 
(1) First, we compute a token-wise \textbf{Evidence Score} via a lightweight attention probe to measure the significance of each incoming view. 
(2) Second, we update a fixed-capacity global \textbf{Memory} that persistently tracks the highest-scoring frames for each individual query token. 
(3) Finally, we perform \textbf{Conditioning-view selection} by allowing the tokens to "vote" for their preferred frames, aggregating these local preferences to select an optimal, bounded-size condition set for the full generation pass.

\textbf{Evidence Score.}
At each chunk, we run a small number of flow/denoising steps of the structure-stage generator on the chunk's conditioning set, using a \emph{frozen} prior $z_0$ sampled once at chunk $0$ and reused for all subsequent chunks. From a fixed cross-attention layer $L$, we extract the cross-attention map in Eq.~\eqref{eq:xattn}. For each non-special query token $q$ and each view $v$, we compute a per-token evidence score $\mathbf{M}_v[q]$ that measures how informative view $v$ is for reconstructing token $q$.

The evidence score should capture two complementary properties: the total amount of attention assigned to view $v$ and how spatially concentrated that attention is within the view. We therefore define a normalized attention distribution over the $P$ patch tokens of view $v$:
\begin{equation}
\mathbf{H}_v[q] =
-\frac{1}{\log(P)}
\sum_{i=1}^{P}
\tilde{\mathbf{A}}_v[q,i]\log \tilde{\mathbf{A}}_v[q,i],
\qquad
\tilde{\mathbf{A}}_v[q,i]
=
\frac{\mathbf{A}_v[q,i]}
{\sum_{j=1}^{P}\mathbf{A}_v[q,j]} .
\end{equation}
Here, $\mathbf{H}_v[q]\in[0,1]$ is the normalized entropy of the attention distribution, where lower entropy indicates that the query token attends to a more localized set of patches.
We then define the evidence score as
\begin{equation}
\mathbf{M}_v[q]
=
\left(1+
\beta
\left(\sum_{i=1}^{P}\mathbf{A}_v[q,i] -
\frac{1}{N_c}\sum_{u=1}^{N_c}\sum_{i=1}^{P}\mathbf{A}_u[q,i]\right)
\right)\cdot
\left(1-\mathbf{H}_v[q]\right),
\end{equation}
where $N_c$ is the number of views in the current chunk and $\beta=8$ is a scaling hyperparameter. The first factor measures the relative cross-attention mass assigned to view $v$ compared with the chunk average, while the second factor rewards spatially concentrated attention.

This design follows a scale--concentration decomposition. A view provides strong evidence for token $q$ only when it receives above-average attention mass and that attention is focused on a compact, spatially specific set of patches. In contrast, high-entropy attention spread over many patches is treated as weak evidence, even if the total attention mass is large, since it may correspond to diffuse background response or non-specific visual context.
Reusing the frozen prior $z_0$ is also intentional. It makes evidence scores comparable across chunks: changes in the attention maps are driven by differences in the conditioning views rather than by a fresh noise sample at each chunk.

\textbf{Memory.}
The cross-chunk memory is represented by two matrices $\mathbf{M}, \mathbf{F} \in Q \times D$, where $Q$ is the number of query tokens, and $D$ is the number of frames to retain, i.e., memory depth.
$\mathbf{M} \in \mathbb{R}^{Q \times D}$ is the \emph{evidence memory}, which stores the $D$ token-wise highest evidence scores over all observed views; $\mathbf{F} \in \mathbb{R}^{Q \times D}$ is the \emph{frame-index memory}, which stores their corresponding global frame indices. 
During the streaming process, each token's top-$D$ list is updated by merging in the new candidates from newly arriving frames.
Frames that never enter any token's list are discarded immediately. 

\textbf{Conditioning-view Selection.}
At a certain chunk, the runtime view set $\mathcal{V}^{\star}$ selected for the full forward pass is obtained by aggregating the per-token top-$D$ lists into a per-frame \emph{token-ownership count}, then taking the \textbf{top-$K$} frames by that count. 
Concretely, we define the ownership count of frame $f_r$ as the number of (token, rank) slots it occupies anywhere in $\mathbf{F}$:
\begin{equation}
\label{eq:ownership}
n_{f_r} \;=\; \big| \{ (q, j) : \mathbf{F}[q, j] = {f_r},\ 1 \le q \le Q,\ 1 \le j \le D \} \big|,
\end{equation}
and select the $K$ frames with the largest counts for $\mathcal{V}^{\star}$.
Notably, there are two distinct ranks $D$ and $K$. $D$ is the \emph{memory depth}: how many candidate frames each token retains in its sorted list; $K$ is the \emph{bundle size}: how many frames the downstream generator consumes per forward pass. 
Here, $D$ controls how robust the evidence score computation is.

\subsection{Evidence-Based Multi-Generation}
\label{sec:method:generation}

\textbf{Generation via multi-view flow-matching fusion.}
The selected views $\mathcal{V}^{\star}$ are passed to the
generator $f_\theta$, which performs multi-view-conditioned diffusion: at each step $t$, the generator computes a velocity $V_\theta(z_t, v)$ for each $v \in \mathcal{V}^{\star}$ and fuses the per-view scores into a single update on the shared latent $z_t$. Following the standard multi-diffusion fusion rule, the fused velocity at each query token $q$ is a view-weighted average:
\begin{equation}
\label{eq:fusion}
\bar{V}_\theta(z_t)[q] \;=\; \sum_{v \in \mathcal{V}^{\star}} \overline{M}_v[q]\, V_\theta(z_t, v)[q],
\qquad
\sum_{v \in \mathcal{V}^{\star}} \overline{M}_v[q] = 1,
\end{equation}
where $\overline{M}_v[q]$ is the normalized per-token, per-view evidence score. 
The full forward pass is $\hat{y} = f_\theta(\mathcal{V}^{\star}; z_k)$ with $z_k$ drawn freely per chunk.


\begin{algorithm}[t]
\caption{Streaming inference with Adaptive Evidential Memory.}
\label{alg:vcc}
\begin{algorithmic}[1]
\Require
  frozen generator $f_\theta$, stream chunks $\{\mathcal{C}_k\}_{k\ge 0}$, attention layer $L$,
  query-token range $[a,b]$, memory depth $D$, bundle size $K$.
\State Sample $z_0 \sim \mathcal{N}(0,I)$ \Comment{frozen prior for evidence probing}
\State Initialize evidence memory $\mathcal{M} \gets \mathbf{0}^{Q \times D}$
\State Initialize frame-index memory $\mathcal{F} \gets \mathbf{0}^{Q \times D}$
\For{$k = 0,1,2,\ldots$}
    \State Receive current chunk $\mathcal{C}_k$
    \State Run warmup flow/denoising steps of $f_\theta$ on $\mathcal{C}_k$ using the frozen prior $z_0$
    \State Extract cross-attention maps $\{\mathbf{A}_v\}_{v\in\mathcal{C}_k}$ from layer $L$ via Eq.~\ref{eq:xattn}
    \State Compute per-view evidence scores $\mathbf{m}_v \in [0,1]^Q$ over patch tokens for each $v\in\mathcal{C}_k$
    \State Update $(\mathcal{M},\mathcal{F})$ by row-wise top-$D$ merging over query tokens
    \State Compute frame ownership counts $\{n_f\}$ from $\mathcal{F}$ via Eq.~\ref{eq:ownership}
    \State Select the conditioning bundle $\mathcal{V}^{\star}_k$ from the top-$K$ frames
    \State Sample a fresh generation prior $z_k \sim \mathcal{N}(0,I)$
    \State Generate $\hat{y}_k \gets f_\theta(\mathcal{V}^{\star}_k; z_k)$ with evidence-weighted fusion in Eq.~\ref{eq:fusion}
    \State \Return $\hat{y}_k$
\EndFor
\end{algorithmic}
\end{algorithm}

\textbf{Algorithm and properties.}
Algorithm~\ref{alg:vcc} summarizes the per-chunk procedure. The method involves three integer hyperparameters ($L$, $D$, $K$), a fixed prior $z_0$, and no learned parameters. Processing each chunk requires a single warmup forward pass to extract cross-attention at layer $L$, followed by memory updates, and a full forward pass of $f_\theta$ using the selected conditioning frame bundle.

\textbf{Summary.} The Adaptive Evidential Memory has two structural properties that clearly distinguish it from latent-transport schemes. First, its cross-chunk memory footprint is bounded by at most $2\times Q\times D$ scalars, and is therefore constant with respect to the stream length $T$. Second, for every query token $q$ and rank $j$, the cached evidence score $\mathbf{M}[q,j]$ is monotonically non-decreasing over time, since each update retains only the top-$D$ entries from the union of the previous cache and the newly arrived candidates. As a result, the conditioning bundle selected for 3D generation can only stay the same or improve in terms of evidence quality as streaming progresses, while the memory cost remains fixed.
\section{Experiment}

We evaluate \method{} on long-stream 3D generation, where the model receives a sequence of continuously arriving posed images and must maintain a coherent 3D representation over time. Unlike standard multi-view reconstruction, this setting stresses two properties simultaneously: the method must exploit newly observed views to improve geometry and appearance, while preserving long-range consistency without reprocessing the entire stream. Our experiments are designed to answer three questions: (i) whether \method{} improves streaming 3D generation quality over single-view and multi-view baselines, (ii) whether token-wise evidential memory provides better long-range consistency than existing streaming alternatives, and (iii) how memory size and view-selection strategy affect performance.

\begin{figure*}[t]
\begin{center}
\includegraphics[width=\linewidth]{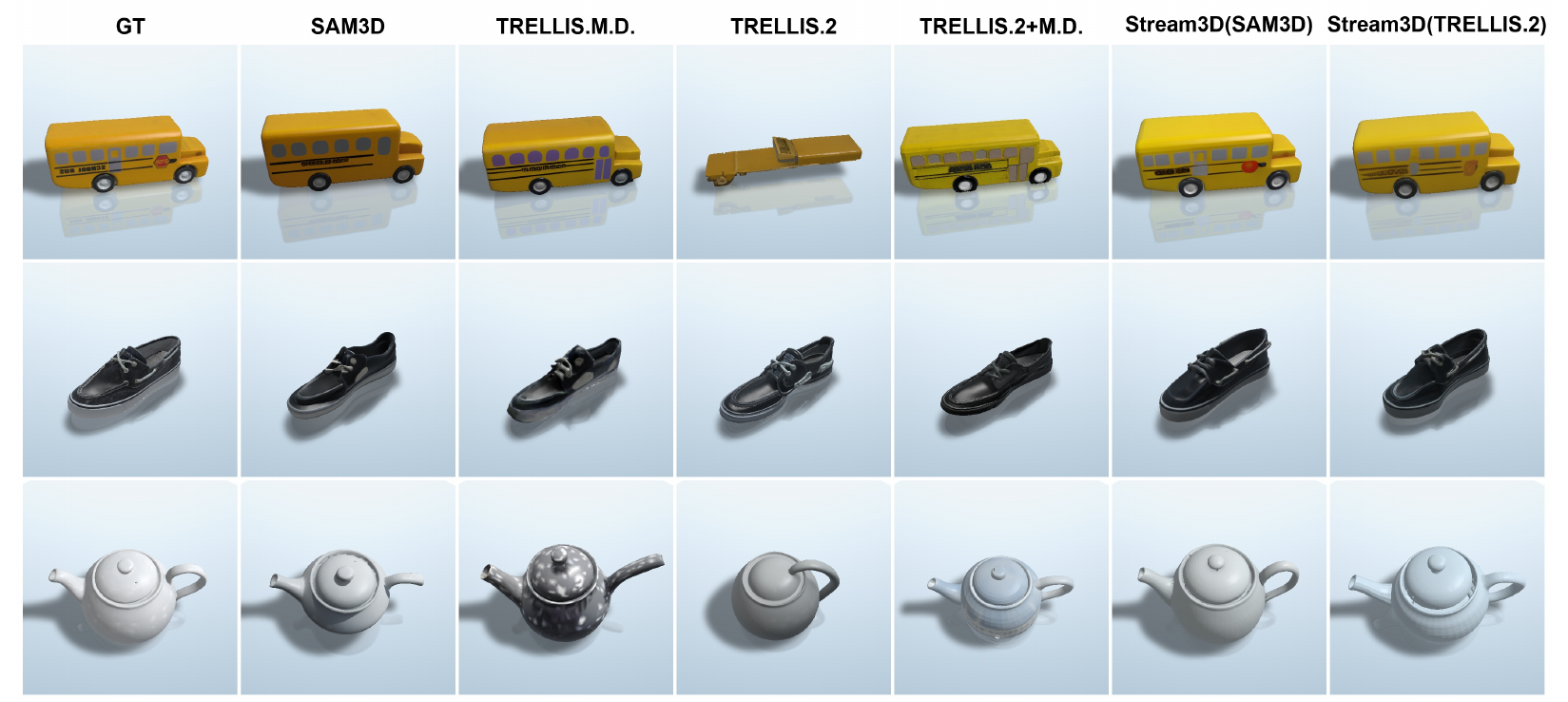}
\end{center}
\vspace{-0.2cm}
\caption{\textbf{Qualitative results on GSO and NAVI.}  \method{} produces more consistent and geometrically faithful 3D generations than single-view and multi-view diffusion baselines.
}
\vspace{-0.25cm}
\label{fig:main_fig}
\end{figure*}

\begin{table}[t]
\centering
\caption{\textbf{Main results on GSO and NAVI.} \method{} consistently improves geometry and appearance over single-view and multi-view generation baselines.}
\label{tab:main_results}
\resizebox{\linewidth}{!}{
\begin{tabular}{llccc cccc}
\toprule
\multirow{2}{*}{Data} & \multirow{2}{*}{Method}
& \multicolumn{3}{c}{Geometry}
& \multicolumn{4}{c}{Appearance} \\
\cmidrule(lr){3-5} \cmidrule(lr){6-9}
& & 
\cellcolor{col2} CD$\downarrow$ &
\cellcolor{col1} IOU$\uparrow$ &
\cellcolor{col2} PFID$\downarrow$ &
\cellcolor{col1} PSNR$\uparrow$ &
\cellcolor{col1} SSIM$\uparrow$ &
\cellcolor{col2} LPIPS$\downarrow$ &
\cellcolor{col2} Image FID$\downarrow$ \\
\midrule
\multirow{6}{*}{GSO}
& StreamVGGT & 0.064 & 0.549 & 75.520 & 10.800 & 0.787 & 0.258 & 189.180 \\
& EscherNet + NeuS & 0.063 & 0.664 & 49.570  & 14.790 & 0.840 & 0.151 & 120.990  \\
& TRELLIS.2           & 0.146 & 0.433 & 140.513 & 12.281 & 0.830 & 0.201 & 141.006 \\
& TRELLIS+M.D.        & 0.053 & 0.750 & 44.515  & 15.933 & 0.854 & 0.149 & \textbf{60.091} \\
& TRELLIS.2+M.D.      & 0.093 & 0.648 & 87.720  & 14.082 & 0.839 & 0.176 & 98.255 \\
& SAM3D               & 0.094 & 0.664 & 71.263  & 14.178 & 0.848 & 0.178 & 105.197 \\
& Stream3D(TRELLIS.2) & 0.087 & 0.692 & 77.730  & 14.213 & 0.845 & 0.169 & 83.279 \\
& Stream3D(SAM3D)     & \textbf{0.048} & \textbf{0.775} & \textbf{40.641} & \textbf{16.145} & \textbf{0.866} & \textbf{0.139} & 66.711 \\
\midrule
\multirow{6}{*}{NAVI} 
& TRELLIS.2           & 0.071 & 0.818 & 42.045 & 20.362 & 0.934 & 0.075 & 88.067 \\
& TRELLIS+M.D.        & 0.055 & 0.810 & 44.297 & 20.426 & 0.935 & 0.078 & 87.654 \\
& TRELLIS.2+M.D.      & 0.066 & 0.791 & 50.839 & 19.914 & 0.934 & 0.078 & 89.755 \\
& SAM3D               & 0.068 & 0.799 & 47.989 & 19.445 & 0.930 & 0.086 & 89.593 \\
& Stream3D(TRELLIS.2) & 0.045 & \textbf{0.833} & \textbf{40.431} & \textbf{20.880} & 0.936 & \textbf{0.072} & 85.546 \\
& Stream3D(SAM3D)     & \textbf{0.043} & 0.825 & 42.151 & 20.720 & \textbf{0.939} & 0.073 & \textbf{81.173} \\
\bottomrule
\end{tabular}
}
\vspace{-0.5cm}
\end{table}

\subsection{Experimental Setup}

\textbf{Implementation}:
Our experiments are run on NVIDIA H100 GPU. We use SAM3D with its standard settings as the underlying generation backbone. 
We adopt Depth Anything 3~\cite{lin2025depth} to estimate the camera pose and depth of initial input to get a point map, which are fed into SAM 3D for 3D generation.
We set $K=8$ and $D=5$ for efficiency, and provide ablation studies to justify this choice. In all experiments, we evaluate multi-view generation models using streams of length 100.

\textbf{Dataset}: We evaluate on two complementary benchmarks. The first is the GSO benchmark~\cite{downs2022google}, which contains high-quality scanned objects with ground-truth 3D assets. Following prior multi-view generation and reconstruction protocols, we render posed image streams from each object and evaluate both novel-view appearance and geometric accuracy. This controlled setting allows us to measure whether the generated 3D content remains faithful to the underlying object as the stream length increases.
The second benchmark is NAVI~\cite{jampani2023navi}, which provides more complex object-centric view sequences with diverse camera trajectories and challenging viewpoint variation. 

\textbf{Baseline}: We compare \method{} with representative single-view, multi-view, and streaming baselines. Single-view baselines, including SAM3D, and TRELLIS, generate 3D content from individual observations and therefore provide a lower bound on streaming consistency. EscherNet~\cite{kong2024eschernet} serves as a strong multi-view baseline that benefits from multiple posed observations but is not designed for unbounded streaming inputs. 
StreamVGGT~\cite{zhuo2026streaming} serves as a reconstruction baseline with streaming input. 
We also compare against TRELLIS+M.D.~\cite{trellis} and TRELLIS.2+M.D.~\cite{trellis2} which aggregate multiple views through full-context or multi-window diffusion but becomes expensive as the number of frames grows.

\textbf{Metrics.}
We report appearance and geometry metrics. For appearance, we use Peak Signal-to-Noise Ratio (PSNR), Structural Similarity Index Measure (SSIM), Learned Perceptual Image Patch Similarity (LPIPS) and Fréchet Inception Distance (FID) on held-out novel views. These metrics evaluate whether the generated representation renders images that are both photometrically accurate and perceptually faithful. For geometry, we report Patch Fréchet Inception Distance (PFID), Chamfer Distance (CD) and Intersection over Union (IOU). These metrics measure absolute depth quality, relative geometric consistency, and fine-grained 3D accuracy.

\begin{figure*}[t]
\begin{center}
\includegraphics[width=\linewidth]{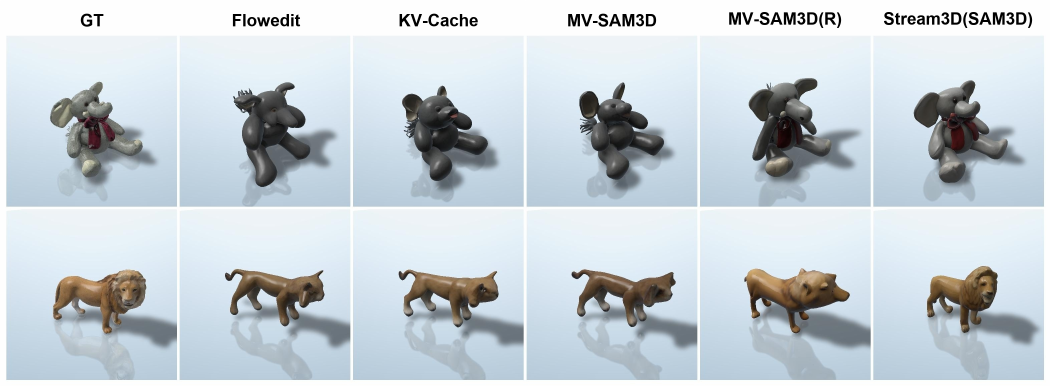}
\end{center}
\vspace{-0.2cm}
\caption{\textbf{Qualitative result of our Ablation studies.}
FlowEdit denotes SAM3D with FlowEdit, and KV-Cache denotes SAM3D with KV-cache reuse. MV-SAM3D denotes MV-SAM3D applied to the last input chunk, while MV-SAM3D(R) denotes MV-SAM3D with $K$ randomly selected views.
}
\label{fig:ablation}
\vspace{-0.5cm}
\end{figure*}

\subsection{Main Results}

Table~\ref{tab:main_results} reports results on GSO and NAVI, and Fig.~\ref{fig:main_fig} shows qualitative results on GSO. Across both datasets, Stream3D achieves the strongest overall performance on appearance and geometry metrics. On GSO, Stream3D(SAM3D) obtains the best CD, IoU, P-FID, PSNR, SSIM, and LPIPS, while also remaining competitive on Image FID. On NAVI, Stream3D achieves the best CD and Image FID, with consistently strong appearance scores. These results indicate that the improvement is not limited to rendering quality but also reflects better 3D structure.
We draw two conclusions. (1) Compared with single-view baselines such as SAM3D, TRELLIS.2, and TRELLIS.2, Stream3D substantially improves both appearance and geometry. This confirms that streaming observations provide useful information that cannot be recovered from a single image alone. Single-view methods often generate plausible visible regions but hallucinate unobserved geometry inconsistently. (2) Compared with multi-view baselines such as TRELLIS+M.D. and TRELLIS.2+M.D., Stream3D shows stronger long-stream behavior. Multi-view diffusion improves consistency with a bounded view set, but does not provide an explicit mechanism for compact long-range memory. Stream3D addresses this by retaining token-level evidence and selecting a bounded conditioning bundle, maintaining high reconstruction quality while keeping memory usage constant.

\begin{table}[t]
\centering
\caption{\textbf{Ablation studies of different streaming strategies.}}
\label{tab:ab_results}
\resizebox{\linewidth}{!}{
\begin{tabular}{llccccccc}
\toprule
\multirow{2}{*}{Data} & \multirow{2}{*}{Method} & \multicolumn{3}{c}{Geometry} & \multicolumn{4}{c}{Appearance} \\
\cmidrule(lr){3-5} \cmidrule(lr){6-9}
& & \cellcolor{col2} CD$\downarrow$ & \cellcolor{col1} IOU$\uparrow$ & \cellcolor{col2} PFID$\downarrow$ & \cellcolor{col1} PSNR$\uparrow$ & \cellcolor{col1} SSIM$\uparrow$ & \cellcolor{col2} LPIPS$\downarrow$ & \cellcolor{col2} Image FID$\downarrow$ \\
\midrule
\multirow{6}{*}{GSO} 
& SAM3D + FlowEdit & 0.090 & 0.668 & 76.445 & 14.343 & 0.850 & 0.178 & 98.643 \\
& SAM3D + KV-Cache & 0.084 & 0.682 & 67.292 & 14.482 & 0.852 & 0.171 & 83.353 \\
& MV-SAM3D + Last Chunk & 0.113 & 0.630 & 86.044 & 13.906 & 0.848 & 0.184 & 104.228 \\
& MV-SAM3D with K random views & 0.064 & 0.676 & 68.534 & 14.828 & 0.859 & 0.156 & 83.039 \\
& MV-SAM3D + Visibility & 0.063 & 0.716 & 58.881 & \textit{15.05} & \textit{0.855} & \textit{0.163} & 81.219 \\
& MV-SAM3D + VLM & 0.059 & 0.710 & 47.990 & \textit{15.23} & \textit{0.855} & \textit{0.157} & 78.530 \\
& MV-SAM3D + Stream3D & 0.053 & 0.760 & 47.260 & 15.892 & 0.864 & 0.1449 & 77.750  \\
& Ours (Fast) & 0.051 & 0.759 & 43.940 & 15.510 & 0.861 & 0.152 & 80.610 \\
& Ours & \textbf{0.048} & \textbf{0.775} & \textbf{40.641} & \textbf{16.145} & \textbf{0.866} & \textbf{0.139} & \textbf{66.711} \\
\bottomrule
\end{tabular}}
\vspace{-0.5cm}
\end{table}

\subsection{Streaming Baseline Comparison}

Table~\ref{tab:ab_results} compares Stream3D with several streaming alternatives, and Fig.~\ref{fig:ablation} reports qualitative results. 
(1) We also compare against cache- and transport-based streaming baselines. KV-cache reuse efficiently carries information across chunks, but cached states are not explicitly filtered by geometric reliability and can accumulate stale or ambiguous evidence. FlowEdit improves short-range consistency through local latent editing, but it operates on fixed-size chunks and loses long-range observations outside the editing window. In contrast, Stream3D does not transport all past latent states. It keeps a compact persistent memory and updates it according to token-level evidence.
(2) We first compare against MV-SAM3D-style fixed-view selection to test whether a compact set of representative frames can replace streaming memory. Random view sampling is unstable because it may discard observations that are critical for reconstructing specific regions. Visibility- and VLM-based selection improve over random sampling, but they still operate at the frame level. This reveals the limitation of global view selection: different spatial tokens may require evidence from different historical views. In contrast, Stream3D maintains token-level evidential memory, allowing each spatial region to retain the observations that best support its generation.
(3) Finally, MV-SAM3D+Stream3D uses our selected views with MV-SAM3D fusion. Its improvement over other MV-SAM3D variants shows that evidential selection is useful by itself, while the remaining gap to Stream3D shows that evidence-weighted fusion also matters. 
(4) To further demonstrate the generalizability of Stream3D, we also implement it on the shortcut version of SAM3D, denoted as Ours (Fast). The results show that even with the shortcut sampler, Ours (Fast) still achieves strong performance.
Overall, Stream3D provides a more flexible and scalable mechanism for long-stream generation, preserving global consistency without storing or jointly fusing all past frames.

\subsection{Ablation Studies}
We conduct ablation studies in Tab.~\ref{tab:full_ablation_results} to analyze the effects of evidence stage, evidence normalization, memory depth, and TopK view selection.
First, we study which generation stage should be used for evidence computation. Following~\cite{chen2025sam}, we denote the first generation stage as SS and the second as SLAT. Deeper stages produce more reliable geometry-aware evidence: SS stabilizes around stage 9 and SLAT around stage 6. Among the tested combinations, SS9-SLAT6 achieves strong geometry and appearance performance, indicating that later transformer features provide better multi-view correspondence for streaming view selection.
Second, we compare evidence scoring variants. Our normalized Evidence score outperforms entropy-only scoring (ENT) and the unnormalized variant Evidence(N). ENT provides weaker geometric alignment, while Evidence(N) is less stable and leads to higher perceptual errors. This shows that normalization is important for balancing confidence across candidate views and avoiding dominance by uncalibrated attention magnitudes.
Third, we evaluate memory depth $D$. Increasing $D$ from 1 to 3 or 5 improves long-range consistency by retaining multiple high-evidence historical views. However, further increasing $D$ to 7 does not consistently help, likely because weaker or stale evidence can enter the memory. Overall, $D=5$ provides a stable trade-off between reconstruction quality, appearance, and memory cost.
Finally, we analyze TopK view selection. Using too few views ($K=4$) reduces multi-view coverage and degrades performance. Once enough informative views are selected, performance becomes relatively stable: increasing $K$ from 8 to 16 yields only moderate gains. We therefore use $K=8$ as a practical default, balancing quality and efficiency.

\begin{table}[t]
\centering
\caption{\textbf{Ablation studies on module selection and hyperparameters.} ``Stage'' denotes the step at which evidence scores are computed. ``E'' denotes the evidence computation strategy. ``D'' denotes the memory depth. ``K'' denotes the number of selected Top-K views. Our setting is \textcolor{red}{highlighted}.}
\label{tab:full_ablation_results}
\resizebox{\linewidth}{!}{
\begin{tabular}{llllccccccc}
\toprule
\multirow{2}{*}{Stage} & \multirow{2}{*}{E} & \multirow{2}{*}{D} & \multirow{2}{*}{K}
& \multicolumn{3}{c}{Geometry}
& \multicolumn{4}{c}{Appearance} \\
\cmidrule(lr){5-7} \cmidrule(lr){8-11}
& & & &
\cellcolor{col2} CD-L2$\downarrow$ &
\cellcolor{col1} $\alpha$-IoU$\uparrow$ &
\cellcolor{col2} P-FID$\downarrow$ &
\cellcolor{col1} PSNR$\uparrow$ &
\cellcolor{col1} SSIM$\uparrow$ &
\cellcolor{col2} LPIPS$\downarrow$ &
\cellcolor{col2} Img-FID$\downarrow$ \\
\midrule
SS6-SLAT3 & Evi  & 5 & 8  & 0.057 & 0.744 & 48.62 & 15.577 & 0.859 & 0.1531 & 72.16 \\
SS6-SLAT6 & Evi  & 5 & 8  & 0.057 & 0.742 & 49.09 & 15.536 & 0.858 & 0.1528 & 70.97 \\
SS9-SLAT3 & Evi  & 5 & 8  & \underline{0.048} & 0.775 & 41.36 & 16.136 & \underline{0.866} & 0.1406 & 68.28 \\
\textcolor{red}{SS9-SLAT6} & \textcolor{red}{Evi}  & \textcolor{red}{5}  & \textcolor{red}{8}  & \underline{0.048} & 0.775 & 40.64 & 16.145 & \underline{0.866} & \underline{0.1396} & 66.71\\
SS9-SLAT9 & Evi  & 5 & 8  & \underline{0.048} & \underline{0.776} & 41.05 & 16.119 & 0.865 & 0.1403 & 67.88 \\
\midrule
SS9-SLAT6 & Evi(U) & 5 & 8  & 0.056 & 0.749 & 44.98 & 15.730 & 0.863 & 0.1486 & 74.08 \\
SS9-SLAT6 & ENT & 5 & 8  & 0.053 & 0.760 & 47.26 & 15.892 & 0.864 & 0.1449 & 67.75 \\
\midrule
SS9-SLAT6 & Evi & 1 & 8 & 0.051 & 0.769 & 41.15 & 15.952 & \underline{0.866} & 0.1406 & 67.93 \\
SS9-SLAT6 & Evi & 3 & 8 & 0.049 & 0.773 & \textbf{39.47} & 16.067 & \underline{0.866} & 0.1415 & 66.45 \\
SS9-SLAT6 & Evi  & 7 & 8  & 0.049 & 0.772 & 40.19 & 16.040 & 0.863 & 0.1425 & 67.50 \\
\midrule
SS9-SLAT6 & Evi  & 5 & 4  & 0.055 & 0.755 & 45.47 & 15.741 & 0.863 & 0.1478 & 70.20 \\
SS9-SLAT6 & Evi  & 5 & 12 & \underline{0.048} & 0.775 & \underline{40.12} & \underline{16.167} & 0.865 & 0.1411 & \underline{66.29} \\
SS9-SLAT6 & Evi  & 5 & 16 & \textbf{0.048} & \textbf{0.779} & 41.48 & \textbf{16.169} & \textbf{0.867} & \textbf{0.1393} & \textbf{65.69} \\
\bottomrule
\end{tabular}}
\vspace{-0.25cm}
\end{table}

\begin{table}[t]
\centering
\caption{\textbf{Efficiency.} Full-length efficiency on a clean single-H100 setting.}
\label{tab:runtime}
\resizebox{\linewidth}{!}{
\begin{tabular}{lccccc}
\toprule
Model & Warmup / chunk (s) & Generation total (s) & Total / sequence (s) & Per frame (s) & Peak GPU (GB) \\
\midrule
Stream3D & 6.9 & 955 & 1062 & 10.6 & 18.4 \\
Stream3D-Fast & \textbf{6.7} & \textbf{486} & \textbf{590} & \textbf{5.9} & \textbf{18.3} \\
MV-SAM3D & n/a & 945 & 945 & 9.45 & 18.4 \\
\bottomrule
\end{tabular}
}
\vspace{-0.75cm}
\end{table}

\begin{wraptable}{r}{0.42\textwidth}
\vspace{-0.5em}
\centering
\scriptsize
\renewcommand{\arraystretch}{1.05}
\setlength{\tabcolsep}{12pt}
\caption{\textbf{Evidence memory footprint.}}
\begin{tabular}{ccc}
\toprule
$D$ & Evidence memory & Shape \\
\midrule
1 & 96 KB & $4096 \times 2$ \\
2 & 192 KB & $4096 \times 4$ \\
3 & 288 KB & $4096 \times 6$ \\
4 & 384 KB & $4096 \times 8$ \\
5 & 480 KB & $4096 \times 10$ \\
\bottomrule
\end{tabular}
\label{tab:memory_footprint}
\vspace{-0.8em}
\end{wraptable}
\subsection{Efficiency and Scalability}

Stream3D adds one warmup probe per chunk to extract cross-attention evidence. As shown in Tab.~\ref{tab:runtime}, this probe introduces a fixed overhead, but the dominant cost remains the frozen multi-view generation backbone. Under the same view budget, the generation time of Stream3D is close to MV-SAM3D, indicating that the evidential-memory update does not change the main computational bottleneck.
The persistent memory itself is lightweight. Tab.~\ref{tab:memory_footprint} shows that it stores only the top-$D$ evidence scores and frame indices for each query token, so its size scales as $O(QD)$ rather than $O(T)$. In practice, reconstructing the selected bundle also requires the RGB frames and metadata, such as poses and depths, for frames currently referenced by memory. However, unreferenced frames can be discarded in an online setting, so the persistent cache remains bounded by the retained evidence rather than the full stream length.

\section{Conclusion}

We introduced \method{}, a training-free framework that extends frozen view-conditioned 3D generators to long monocular streams. Instead of transporting latent states across chunks, \method{} uses the generator's cross-attention maps to identify historical views that provide reliable conditioning evidence for each 3D query token. This evidence is stored in a compact Adaptive Evidential Memory and used by Evidence-Based Multi-Generation to select a bounded view bundle for each chunk. As a result, \method{} keeps memory constant with stream length while leaving the original generator unchanged.
Experiments on GSO and NAVI show that \method{} improves both appearance and geometry over single-view generators, multi-view diffusion baselines, and streaming alternatives such as KV-cache reuse and flow-based feature editing. These results support our central claim: streaming 3D generation is better addressed as an evidence selection problem than as a latent transport problem. By preserving token-level evidential views rather than unstable latent states, \method{} provides a scalable path toward coherent long-horizon 3D generation from continuously arriving visual observations. Limitations and acknowledgments are provided in the appendix.

\newpage
\bibliographystyle{abbrv}
{\small
\bibliography{egbib}}
\newpage

\section{Additional Analysis}
\label{app:additional_analysis}

\subsection{Efficiency and Memory Footprint}
\label{app:efficiency_analysis}

Stream3D adds one warmup probe per chunk to extract cross-attention evidence.
Tab.~\ref{tab:bundle_size_runtime} shows that generation time scales similarly with bundle size $K$ for Stream3D and MV-SAM3D. This confirms that increasing $K$ mainly affects the shared multi-view generation pass, while the evidential-memory mechanism adds little extra scaling cost. Stream3D-Fast uses the same memory, selection, and fusion mechanism as Stream3D, but replaces the default stage-1 sampler with the official few-step distilled sampler. It therefore represents a latency--quality trade-off within the same framework rather than a separate method.

\begin{table}[H]
\centering
\caption{Generation-time scaling with bundle size $K$.}
\label{tab:bundle_size_runtime}
\begin{tabular}{lccc}
\toprule
Model & $K=4$ gen total (s) & $K=8$ gen total (s) & $K=8$ / $K=4$ \\
\midrule
Stream3D & 649 & 955 & $1.47\times$ \\
MV-SAM3D & 638 & 945 & $1.48\times$ \\
\bottomrule
\end{tabular}
\end{table}

\subsection{Object-Centric Alignment and Evaluation Protocol}
\label{app:alignment_analysis}

Outputs from object-centric 3D generation methods often live in method-specific canonical coordinate frames. Directly comparing them in their native coordinates would conflate reconstruction quality with arbitrary choices of global pose, scale, and axis convention. We therefore align all method outputs to the same benchmark coordinate system before computing metrics.

The alignment uses the generated source geometry, such as mesh vertices or Gaussian centers, and estimates a global Sim(3) transform to match the ground-truth object geometry. We initialize from multiple candidate orientations to account for common canonical-frame differences and then refine with ICP. This alignment is restricted to a global similarity transform: it does not deform geometry, add missing parts, modify texture, or correct local artifacts.

After automatic alignment, we perform a rendered-view audit in the fixed evaluation cameras. This audit catches gross semantic pose or scale errors that can occur when several candidates have similar Chamfer distances but correspond to different object orientations. If such an error is detected, the remaining ranked global Sim(3) candidates are reviewed, and an alternative is selected only when it visibly resolves the alignment failure. Final metrics are recomputed after the accepted alignment. This makes comparisons across heterogeneous methods more consistent while preserving local reconstruction errors for evaluation.

\begin{table}[H]
\centering
\caption{Ablation on evidence-score fusion variants.}
\label{tab:evidence_score_fusion}
\resizebox{\linewidth}{!}{
\begin{tabular}{lccccccc}
\toprule
Variant & CD-L2 $\downarrow$ & IoU $\uparrow$ & P-FID $\downarrow$ & PSNR $\uparrow$ & SSIM $\uparrow$ & LPIPS $\downarrow$ & Img-FID $\downarrow$ \\
\midrule
\textbf{Product (default)} & \textbf{0.0478} & \textbf{0.775} & \textbf{40.64} & \textbf{16.145} & \textbf{0.866} & \textbf{0.1396} & \textbf{66.71} \\
Sum $\alpha=0.25$ & 0.0505 & 0.7681 & 44.21 & 15.82 & 0.8603 & 0.1481 & 75.88 \\
Sum $\alpha=0.50$ & 0.0527 & 0.7591 & 44.36 & 15.64 & 0.8586 & 0.1500 & 76.61 \\
Sum $\alpha=0.75$ & 0.0545 & 0.7536 & 44.42 & 15.55 & 0.8584 & 0.1521 & 77.67 \\
Evi(N) & 0.056 & 0.749 & 44.98 & 15.730 & 0.863 & 0.1486 & 74.08 \\
Ent & 0.053 & 0.760 & 47.26 & 15.892 & 0.864 & 0.1449 & 67.75 \\
\bottomrule
\end{tabular}
}
\end{table}

\subsection{Evidence-Score Fusion}
\label{app:evidence_score_analysis}

Tab.~\ref{tab:evidence_score_fusion} analyzes different evidence-score fusion variants. Our default product form combines two complementary signals: attention scale and attention concentration. The scale term measures whether a view receives above-average attention mass for a query token, while the concentration term measures whether that attention is spatially localized over image patches.

The ablation shows that both terms are necessary. Attention-mass-only evidence can favor views with diffuse, non-specific attention, while entropy-only evidence can favor sharply peaked but weak responses. Weighted-sum variants improve over some single-component scores but remain less stable than the product form. The product score requires a view to be both strongly attended and spatially focused, making the retained evidence more discriminative for token-level 3D generation.

Overall, these results support three conclusions: Stream3D maintains bounded evidence memory independent of stream length, its runtime overhead is mainly a fixed warmup probe while generation cost is dominated by the frozen backbone, and its product-form evidence score better identifies useful historical views than single-component or weighted-sum alternatives.

\subsection{More Visulization}

\begin{figure*}[t]
\begin{center}
\includegraphics[width=\linewidth]{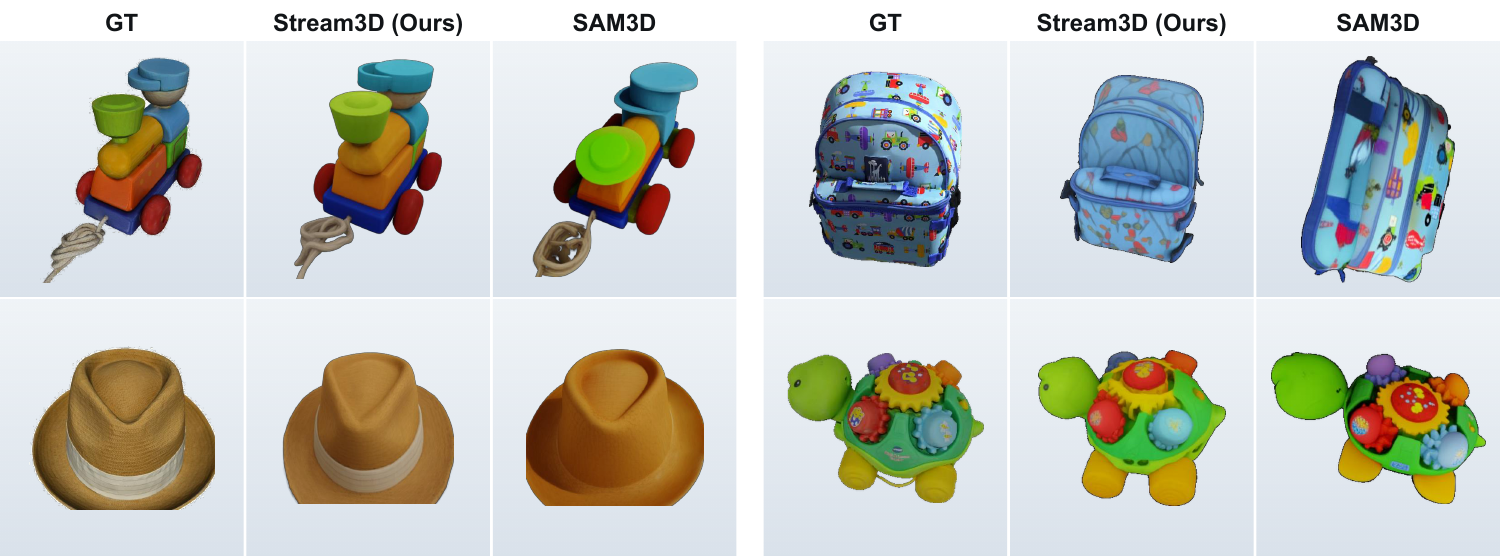}
\end{center}
\caption{\textbf{Long-horizon comparison between SAM3D and \method{}.}
We visualize the generation results of the single-view SAM3D baseline and \method{} as more frames are observed from the input stream. SAM3D processes each observation independently and therefore lacks a mechanism to accumulate evidence across time. In contrast, \method{} maintains an adaptive evidential memory and progressively incorporates informative historical views. As the stream grows, \method{} produces more complete and stable 3D assets, while SAM3D remains limited by the information available from individual frames.}
\label{fig:sam3d_vs_stream3d}
\vspace{-0.5cm}
\end{figure*}

Fig.~\ref{fig:sam3d_vs_stream3d} illustrates the key difference between independent single-view generation and streaming generation. SAM3D produces plausible outputs from individual frames, but its results do not accumulate evidence over time. \method{} instead updates a compact evidential memory as new chunks arrive, allowing later generations to use information from earlier informative views. This leads to progressively more complete reconstructions over long streams.

\begin{figure*}[t]
\begin{center}
\includegraphics[width=\linewidth]{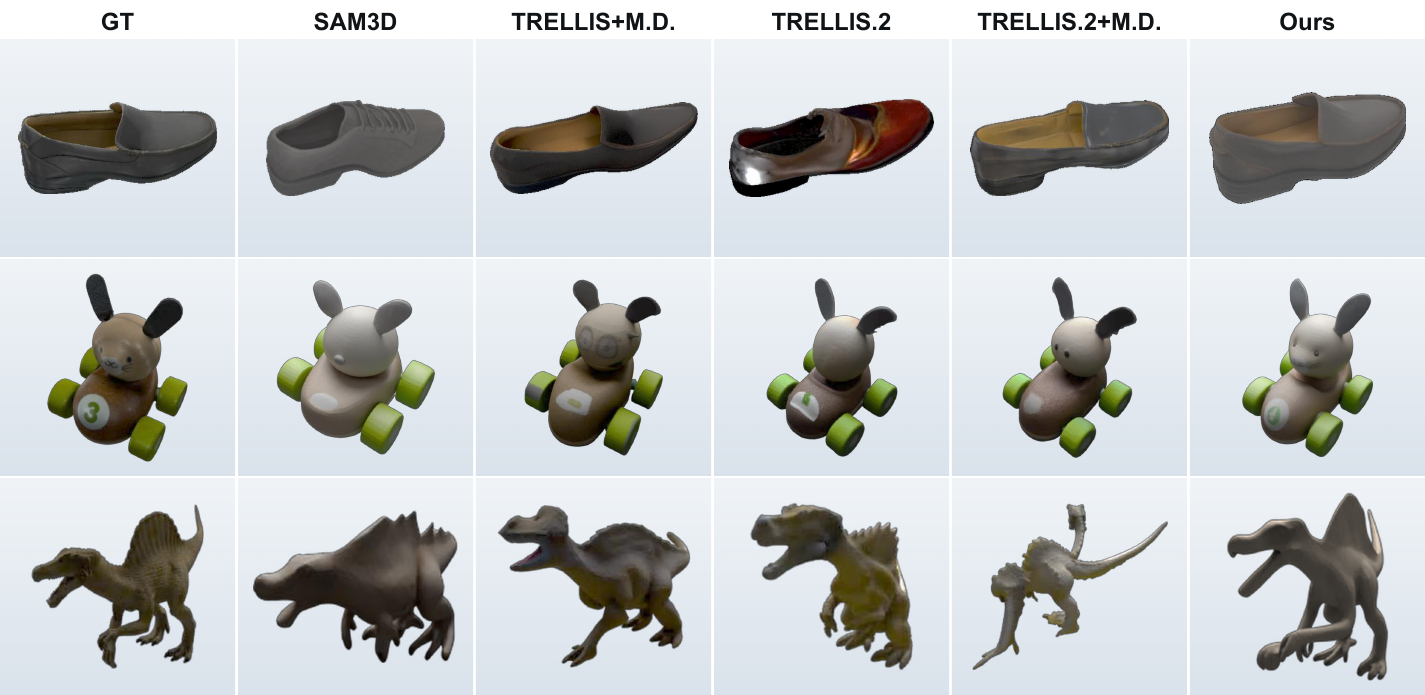}
\end{center}
\vspace{-0.5cm}
\caption{\textbf{Qualitative comparison with single-view and multi-view 3D generation baselines.}
We compare the ground truth with SAM3D, TRELLIS+M.D., TRELLIS.2, TRELLIS.2+M.D., \method{}(SAM3D), and \method{}(TRELLIS.2). Single-view methods recover plausible visible regions but often hallucinate incomplete or inconsistent unseen geometry. Multi-view diffusion improves view consistency with a bounded input set, but can still miss fine geometry or produce artifacts under long-stream observations. \method{} improves both completeness and consistency by selecting informative historical views through token-level evidential memory, and the gains are visible across both SAM3D and TRELLIS.2 backbones.}
\label{fig:main_qualitative}
\vspace{-0.25cm}
\end{figure*}

Fig.~\ref{fig:main_qualitative} further compares \method{} with single-view and multi-view generation baselines. Single-view methods are constrained by partial observation and must hallucinate large unobserved regions. Multi-view diffusion baselines improve consistency by fusing multiple views, but they operate on a bounded view set and lack an explicit long-range memory mechanism. \method{} improves over these baselines by retaining token-level evidence from the stream and selecting the most informative historical observations for generation.

\begin{figure*}[t]
\begin{center}
\includegraphics[width=\linewidth]{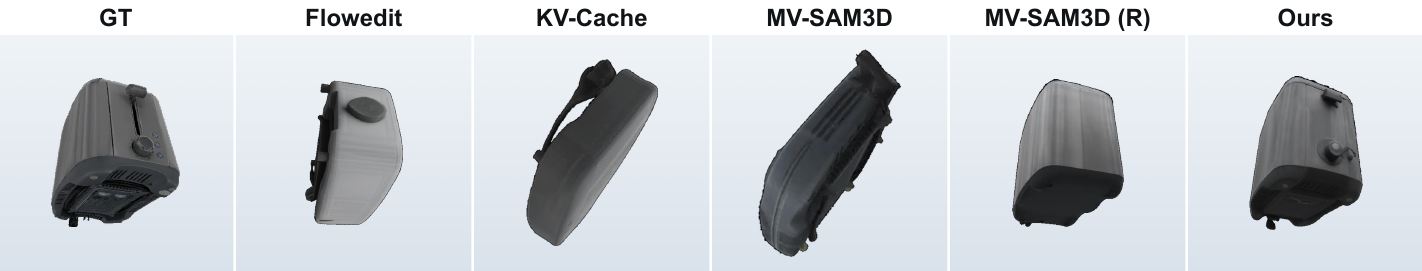}
\end{center}
\vspace{-0.5cm}
\caption{\textbf{Qualitative comparison with streaming baselines.}
We compare the ground truth with FlowEdit, KV-Cache, MV-SAM3D, MV-SAM3D with random view selection, and \method{}(SAM3D). FlowEdit improves short-range consistency but is limited by local latent editing and can lose long-range evidence. KV-Cache carries historical states forward but does not explicitly filter them by geometric reliability. MV-SAM3D variants select a compact set of frames at the view level, which can miss local evidence needed by specific spatial regions. In contrast, \method{} retains token-level evidence over time and selects a bounded conditioning bundle, leading to more complete geometry and more stable appearance under long-stream generation.}
\label{fig:streaming_qualitative}
\vspace{-0.25cm}
\end{figure*}

Fig.~\ref{fig:streaming_qualitative} compares different streaming strategies. Cache- and transport-based methods reuse latent or feature states, but these states can become stale or accumulate errors over long sequences. Frame-level MV-SAM3D variants are more stable, but selecting views globally can discard observations that are important for specific spatial regions. \method{} addresses this by maintaining token-wise evidential memory, which enables different query tokens to rely on different historical views while keeping the conditioning bundle bounded.

\section{Limitations}
The method is built on top of the quality of the underlying pretrained generator. If the base model fails to reconstruct the object from individual views, the proposed evidential memory cannot fully recover the missing geometry or appearance. 

\section{Acknowledgment}
We would like to thank Xihang Yu and Yuzhen Chen for their helpful discussions and support with the robot experiment demos. Their feedback and assistance were valuable to this work.

\end{document}